\pdfoutput=1
\documentclass[11pt]{article}

\usepackage[]{EMNLP2022}

\usepackage{times}
\usepackage{latexsym}
\usepackage{multirow}
\usepackage{graphicx}
\usepackage{booktabs}
 \usepackage{calc}

\usepackage[T1]{fontenc}
\usepackage[utf8]{inputenc}

\usepackage{microtype}

\usepackage{inconsolata}
\usepackage{caption}
\usepackage{subcaption}
\usepackage{float}
\usepackage{placeins}
\usepackage{comment}
\usepackage{xcolor}
\usepackage[symbol]{footmisc}
\renewcommand{\thefootnote}{\fnsymbol{footnote}}

\title{\textsc{EvEntS ReaLM}: Event Reasoning of Entity States via Language Models}

\author{Evangelia Spiliopoulou$^{*\dagger}$ \\
Amazon \\ AWS, AI Labs \\ \texttt{spilieva@amazon.com}\\
\And
Artidoro Pagnoni$^*$\\ Univ. of Washington\\ \texttt{artidoro@cs.}\\ \texttt{washington.edu}
 \And
Yonatan Bisk\\ Carnegie Mellon\\ University \\ \texttt{ybisk@cs.cmu.edu}
 \And
Eduard Hovy\\ 
Carnegie Mellon \\ University\\ \texttt{hovy@cs.cmu.edu}}
\begin{document}
\maketitle
\let\thefootnote\relax\footnote{\textsuperscript{*} Equal contribution.}
\let\thefootnote\relax\footnote{\textsuperscript{$\dagger$} Work completed while at Carnegie Mellon University, before joining AWS AI Labs.}
\thefootnote\relax\footnote{\textsuperscript{1}  \url{https://github.com/spilioeve/eventsrealm}}

\begin{abstract}
This paper investigates models of event implications. Specifically, how well models predict entity state-changes, by targeting their understanding of physical attributes. Nominally, Large Language models (LLM) have been exposed to procedural knowledge about how objects interact, yet our benchmarking shows they fail to \textit{reason} about the world. Conversely, we also demonstrate that existing approaches often misrepresent the surprising abilities of LLMs via improper task encodings and that proper model prompting can dramatically improve performance of reported baseline results across multiple tasks. In particular, our results indicate that our prompting technique is especially useful for unseen attributes (out-of-domain) or when only limited data is available.$^1$

\end{abstract}

\section{Introduction}

Modeling the effect of actions on entities (\textit{event implications}) is a fundamental problem in AI spanning computer vision, cognitive science and natural language understanding. Most commonly referred to as the Frame Problem \cite{mccarthy1981some}, early solutions relied on a set of handcrafted rules and logical statements to model event implications. However, such methods require substantial manual effort and fail to generalize. More recently, modeling event implications has reemerged under the guise of common sense reasoning within NLP \cite{sap2019atomic,bisk2020piqa,talmor-etal-2019-commonsenseqa} and action anticipation in Computer Vision \cite{Damen2018EPICKITCHENS,bakhtin2019phyre}.

Predicting event implications is a particularly difficult problem due to the complex nature of language and implicit knowledge required to answer such questions. For example, if we are given the sentence \textit{the mug fell on the floor} and we want to determine whether \textit{the mug is broken}, we need to know of several facts such as the material of the mug, the fragility of ceramics, the hardness of the floor, etc.\ and also how to combine these facts together to \textbf{reason} whether the mug will break or not. None of this knowledge is explicitly stated, instead being classified as \textit{common sense knowledge}, and is traditionally acquired from observations or interactions with objects and the environment.%, and previous communication. 

\begin{figure}
\includegraphics[width=\columnwidth]{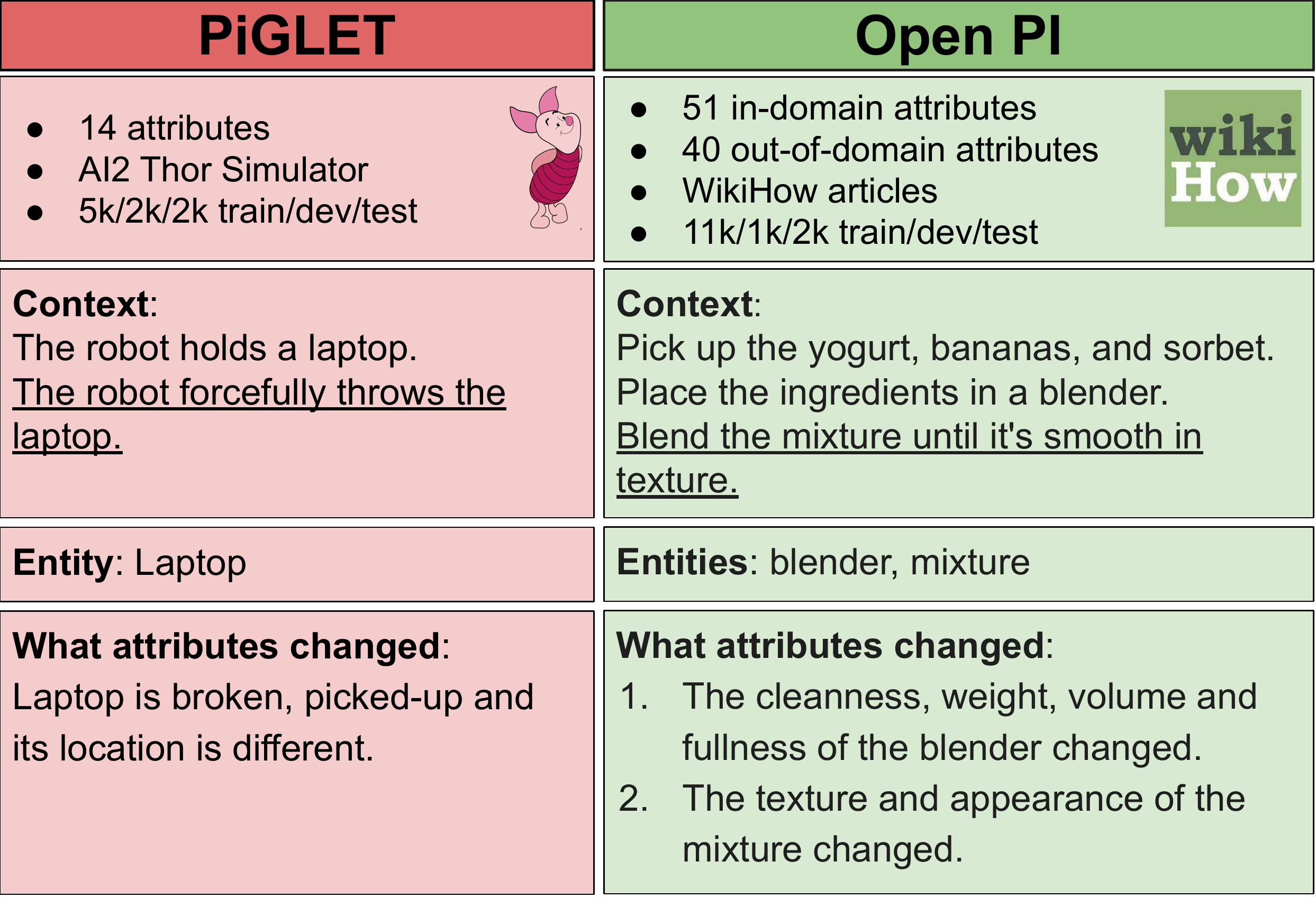}
\caption{We use the PiGLET and OpenPI datasets to probe if LLMs contain the necessary grounded and world knowledge to reason about event implications.}
\label{fig:examples}
\end{figure}

Core to this line of work is the assumption that events can be learned via language, not depending on other forms of perception. To explore the utility of other modalities and interaction, 
\citet{zellers-etal-2021-piglet} train a language model to predict physical changes in a virtual environment.  While intuitively necessary \cite{bisk-etal-2020-experience}, in this work we show that the purported limitations of language-only models are not always well founded. Key to their success (or failure) are (1) How we use the language models, and (2) The difficulty of the task domain and dataset.   

We find that the difficulty of the task is often a stand-in for whether reasoning is required. Others have also noted that despite the tremendous gains in NLU made possible by Large Language models (LLM), they still stumble when reasoning is required \citep{brown2020language}.
%In recent years, we have achieved tremendous advances in several NLU tasks via the use of pre-trained Language Models. Despite their overall good performance, they still struggle in problems requiring reasoning, such as common sense reasoning and reading comprehension \cite{brown2020language}. 
If reasoning can be codified as patterns, we are presented with two new challenges: (1) Can we test pattern acquisition via benchmarking generalization, and (2) How can new patterns or context be provided to the model? 
%This poses several important questions with respect to the limitations of current models and their reasoning abilities, such as whether they are able to generalize in unseen data by applying reasoning patterns. 
The nascent field of ``prompting" \citep{liu2021pre, wei2021finetuned, ouyang2022training} hints at a possible approach for humans to encode novel reasoning patterns for models, however the best structure and 
%Furthermore, even if large language models are able to develop some kind of reasoning mechanism, an additional challenge involves how to exploit what they already know and convey to them sufficient information about the task to enable them extracting the right mechanism. 
%Although techniques like prompting aim to bridge this gap, 
the amount of information to convey via a prompt for a given task still remain open questions.  

%In this work we discuss the problem of event implications as entity change-of-state with respect to their physical attributes. To summarize, our contributions are the following. 
This work makes three contributions to the literature of event implications.
First, we show that language by itself provides enough information to predict event implications in current datasets, without the need of a physical interaction model. Second, we establish the difficulty of the problem and limitations of current models by showing extreme differences in performance  across different datasets: PiGLET \cite{zellers-etal-2021-piglet}, based on a virtual environment, and OpenPI \cite{tandon-etal-2020-dataset}, based on procedural text from WikiHow. Third, we explore how different prompting techniques affect model performance in terms of their information content and model nature. Finally, we discuss the generalization properties of our models to unseen attributes (out-of-domain) and how this shows their ability to extract implicit reasoning mechanisms.

%  To summarize, our contributions are the following:
%  \begin{itemize}
%      \item We show that the performance of physical interaction models is comparable to language-only models and motivate the need for more difficult datasets and domains to quantify the limitations of predicting entity change-of-states from language only.
%      \item We establish the difficulty of the problem by showing extreme performance fluctuation across different types of datasets.
%      \item We propose a prompting technique based on attribute explanations as a means to extract reasoning mechanisms instead of knowledge linked to specific attributes seen during fine-tuning.
%      \item We provide an extensive analysis of performance in out-of-domain attributes across different techniques to share information about the task and probe the generalization properties of the models.  
%  \end{itemize}

\section{Related Work}

Related work in commonsense follows two directions: (1) predict event implications or track entity changes, and (2) use commonsense knowledge about events and their implications as necessary intermediate steps in reasoning. 

% Most recent research in common sense reasoning studies event implications centered around mental states and social interactions \cite{hwang2021comet,sap2019atomic}. Although this line of research highlights the difficulty of predicting cause-effect relations, social scenarios are typically ambiguous and require knowledge of chains of events. For example, in order to answer whether \textit{X gives a gift to Y} implies that \textit{X hugs Y}, we must be aware of the relation of X and Y, their personalities and the social context. On the other hand, event implications as physical change-of-state of entities are, mostly, objective and depend on simple relations that a model could know a priori (e.g., in our example, the material of a mug), allowing us to isolate and study the reasoning abilities of a model.

% Research that directly studies event implications mostly explores causality between social events and emotional states, based on social norm expectations \cite{sap2019atomic,forbes2020social,emelin2020moral, rashkin2018event2mind}. \citet{jiang2021m} study specific linguistic phenomena such as contradiction and negation, while \citet{sap-etal-2019-social} study the role of social biases and the difficulty of automatically predicting implications of social events.   

Research that directly studies event implications mostly explores causality between social events and emotional states, based on social norm expectations \cite{rashkin-etal-2018-event2mind, sap2019atomic,forbes-etal-2020-social,emelin-etal-2021-moral, hwang2021comet}. \citet{jiang-etal-2021-im} study specific linguistic phenomena such as contradiction and negation, while \citet{sap-etal-2019-social} study the role of social biases and predicting implications of social events. Although this line of research highlights the difficulty of predicting cause-effect relations, social scenarios are typically ambiguous and require knowledge of event chains. For example, in order to answer whether \textit{X gives a gift to Y} implies that \textit{X hugs Y}, we must be aware of the relation between X and Y, their personalities, and the social context. On the other hand, event implications as physical changes of state of entities are, mostly, objective and depend on simple relations that a model could know a priori (e.g., the material of a mug), allowing us to isolate and study the reasoning abilities of a model.   

Closer to our task is the prediction of physical implications of events. This problem often takes the form of entity changes in procedural text, such as in cooking recipes \cite{bosselut2018simulating} or WikiHow articles \cite{tandon-etal-2020-dataset}. However, most datasets primarily focus on changes in location compared to other attributes, such as ProPara \cite{dalvi-etal-2018-tracking} and bAbI \cite{weston2016towards}. Modeling approaches in both areas of commonsense explore the generation of explanations in a multi-task setting \cite{dalvi-etal-2019-everything}, the use of external knowledge graph \cite{tandon-etal-2018-reasoning}, and automatic knowledge base construction to keep a representation of the state of the world and generate novel knowledge \cite{bosselut-etal-2019-comet,henaff2017tracking,hwang2021comet}.

The second type of commonsense reasoning includes question answering tasks that assume knowledge of commonsense relations and their implications on the context. This line of work includes short questions, such as OpenBookQA \cite{mihaylov-etal-2018-suit}, CommonSenseQA \cite{talmor-etal-2019-commonsenseqa}, SWAG \cite{zellers-etal-2018-swag} and COPA \cite{roemmele2011choice}, or questions based on a provided document \cite{huang-etal-2019-cosmos} or knowledge base \cite{clark2018think}.

\section{Task and Datasets}
The problem of predicting event implications can be formulated in several ways, with varying levels of difficulty. For example, \citet{tandon-etal-2020-dataset} generate triplets of \textit{entity, attribute, post-state} given some context, while \citet{zellers-etal-2021-piglet} are given an entity, attribute, pre-state, and context, to only predict the \textit{post-state} of the entity.

Our task follows a similar formulation to PiGLET, where the model is given a context (i.e., a small paragraph followed by an action-sentence), an entity of interest and a list of attributes. Then, the model needs to determine whether a change-of-state occurred for the entity with respect to the given list of attributes (see \autoref{fig:examples}). However, unlike \citet{zellers-etal-2021-piglet}, we do not use the pre-state encoding of the entity, instead we assume that the relevant information is better conveyed through the natural language description of the context.

%We use this is because the goal of our task is to evaluate the abilities of our models, the goal of our tall of our experiments rely on evaluating the abilities of a model given only information extracted from language as its input.  

\subsection{PiGLET}

PiGLET \cite{zellers-etal-2021-piglet} consists of encodings of the pre- and post-state of entities as a result of an action. Each instance is accompanied by the \textbf{context}: a natural language description of the pre-state of the entities, followed by a description of the action. PiGLET is a small dataset (5k training examples), which studies entity change-of-state with respect to 14 attributes, caused by 8 distinct events.

PiGLET is a semi-artificial dataset, where the \textit{entity, pre-state, post-state, action} tuple was generated by exploring the virtual environment AI2 Thor \cite{kolve2017ai2}. A natural language context was constructed by human annotators, who were provided with the tuples generated by the virtual environment. This results in simpler concise statements compared to the ambiguous language that humans naturally use to communicate.  

 \subsection{OpenPI}
 \label{data:openpi}
Open PI \cite{tandon-etal-2020-dataset} also studies the change-of-state of entities with respect to physical attributes. However, unlike PiGLET, Open PI is based on articles from WikiHow, containing realistic descriptions of physical changes. The context in this dataset is the entire WikiHow article preceding the action sentence from the article. 

Open PI is a substantially larger dataset, containing an initial set of 51 pre-defined attributes from WordNet \cite{fellbaum2010wordnet}, then augmented by human annotators. Although the total number reaches $\sim$800 unique attributes, the initial 51 attributes cover more than 80\% of instances. Furthermore, the vast majority of the newly introduced attributes appear only once and many of them contain typos or abbreviations. All our models are trained in the initial set of 51 attributes.
%As we explain in Section \ref{openpi_exp}, to test the generalization properties of the models, we treat the initial vocabulary of attributes as in-domain and the new attributes introduced by human annotators as out-domain.

\section{Methodology}

\label{sec:methodology}
Next, we introduce our prompting techniques, which vary with respect to per-instance information content. Each technique is tested with different LLMs and fine-tuning methods. The goal of each prompting mechanism is to show how model performance and generalization vary based on the information conveyed in our queries. Our study focuses on four prompting methods depicted in \autoref{fig:prompts}: zero-prompt, single-attribute, multi-attribute, and a variant of the latter, the $k$-attribute prompt. %The \textbf{zero-prompt}, contains no information about the task and the attributes we want to classify, while the \textbf{single-attribute} and \textbf{multi-attribute} prompts contain information for only one attribute or all attributes, respectively. Finally, the \textbf{k-attribute} prompt is a variant of the multi-attribute model, where we partition the attributes to form multiple queries with varying number of attributes. 

%Our approach builds on the literature demonstrating benefits in using natural language instructions as prompts to convey information about \citep{raffel2020exploring,wei2021finetuned}. 
%In our study, however, we are interested in eliciting the best performance from a LLM on a single task by codifying it as a pattern. To the best of our knowledge we are the first ones to demonstrate advantages and disadvantages with different ways to codify the prompts and use prompting to study LLMs' understanding of event implications for physical attributes.

Our approach builds on literature demonstrating benefits in using prompting to distinguish different tasks, when a model is trained in a multi-task setting \citep{raffel2020exploring,wei2021finetuned}. 
In our study, however, we explore how to use prompts as a medium to convey the task-specific information that a model must know in order to solve the task, similar to how one would ask a human. To the best of our knowledge, we are the first ones to demonstrate advantages and disadvantages of different ways to codify intermediate steps required for reasoning via prompting and use them to study LLMs' understanding of event implications.

% \shruti{is there other work on prompting that uses similar techniques? citations to related work might be useful here}      

%\subsection{Physical Interaction Model}
\subsection{Large Language Models}

We explore three transformer-based language models: an autoregressive, an autoencoder, and a seq-to-seq model. We include models with different architectures to investigate the effect of our prompting strategies across model families. 
Our goal is to use each model in combination with prompts that enhance their individual strengths, based on their pretraining schemes. \\[-5pt]

\noindent \textbf{RoBERTa \cite{liu2019roberta}}: is an autoencoder model widely used in classification tasks. \\[-5pt]

\noindent \textbf{T5 \cite{raffel2020exploring}:} is a seq-to-seq model that has shown excellent performance in multi-tasking by using the task description as a prompt. T5 is used for both text classification and generation.   \\[-5pt]

\noindent \textbf{GPT-3 \cite{brown2020language}:} is an autoregressive model and is primarily used in zero and few-shot settings due to its substantially larger size. GPT-3 is used in language generation and classification, and has shown excellent performance in few-shot settings when queried with appropriate prompts.\\[-5pt]

\begin{figure}
\centering
\includegraphics[width=\columnwidth]{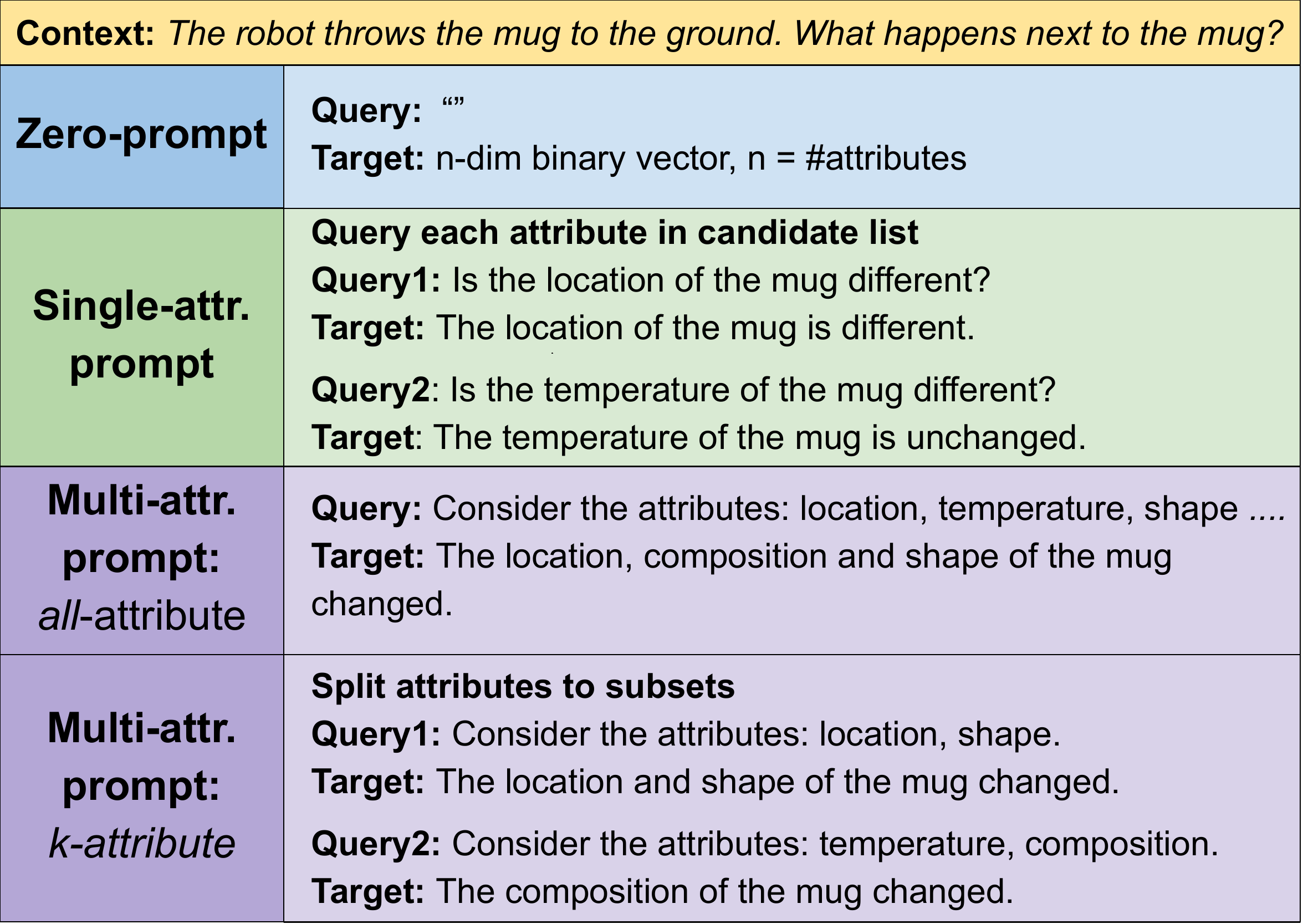}
\caption{Prompting techniques used in our models. Multi-attribute prompt improves performance by learning dependencies among attributes.}
\label{fig:prompts}
\end{figure}

\noindent These backbone models are used with one of the three prompting techniques, as described in the following paragraphs and shown in \autoref{fig:prompts}.

% \textcolor{blue}{This page feels like it might benefit from a small table of example prompts and then shorter descriptions so it's still just one page, but easier to skim}
\subsection{Multi-label Classifier: Zero-prompt}
% Our first model is a multi-label classifier that uses prompts of minimal information about the task to test how well a model can learn which attributes change only based on fine-tuning. The model is given the context, followed by the prompt \textit{Now what happens next to the [entity]?}, where \textit{entity} is replaced with the name of the object whose potential change-of-state we want to predict (see Figure \ref{fig:prompts}). The model predicts a binary vector, where each value corresponds to a change in a specific attribute (multi-label classification). To test this mechanism we used RoBERTa, as it has shown to perform well in classification tasks.

% Through this model, no explicit information is given about the nature of the task or the attributes themselves. We assume that, given enough data, the model should be able to learn which attributes the dimensions of the output vector correspond to, and correctly predict their changes. This model serves as a baseline of how a LLM performs when fine-tuned in a very specific task and does not have the ability to generalize. 

Our baseline model is a multi-label classifier with no explicit information about the nature of the task or the attributes themselves. The model takes the context and the prompt \textit{Now what happens next to the [entity]?} as inputs, and predicts a binary vector, where entries correspond to changes in specific attributes. We test this mechanism with RoBERTa, as it performs well in classification tasks. 

With this model we test the traditional ``finetuning assumption'' that, given enough data, the model can learn the correspondence between attributes and dimensions in the output vector and correctly predict their changes. This model serves as a baseline of how a LLM performs when fine-tuned to a specific task. Crucially, it does not have the ability to generalize to new attributes as the output vector is of fixed size. 

\subsection{LM as Classifier: Single-attribute Prompt}

Our second prompting technique provides information about individual attributes. Via this technique we evaluate whether a model benefits from the verbalization of each attribute, as a means to retain useful information from the context. Unlike the zero-prompt model, this model can be used out-of-domain, with unseen attributes. 

In this setup, we query the model about each individual attribute separately, for every \textit{context-entity} pair, as shown in Figure \ref{fig:prompts}. This mechanism was tested with all three models: RoBERTa (fine-tuned and zero-shot), T5 (fine-tuned) and GPT-3 (few-shot). 

By querying each attribute individually, the model is able to focus only on information related to that specific attribute. This can both benefit and hurt performance, as we show in \autoref{experiments}. On one hand, the model pays more attention to the sentence semantics related to the queried attribute. By using the attribute as a bottleneck, the model learns which aspect of meaning is important in that instance. This is particularly beneficial in limited-data scenarios where generalization is necessary. On the other hand, by querying only a single attribute per instance, the model does not learn correlations across attributes. This weakness becomes more apparent in scenarios with many correlated attributes.   
%By querying each attribute individually, the model is able to: (1) Focus only on information that relates to that specific attribute and (2) Answer queries even for unseen attributes (both in fine-tuned and zero-shot models). 
%However, a drawback compared to the previous approach is that the model sees only one attribute per instance, which means that it cannot understand how  

%We experiment with Roberta and T5 in this setup. We also did experiments with zero-shot Roberta to check modeling improvements via fine-tuning and check the knowledge of the model for certain attributes (location was already very high, at F1 = 55). We also used GPT-3 in a similar fashion as zero-shot. Due to the complex nature of prompts, it was more applicable to use single-attribute than multi-attribute prompts when we cannot finetune TODO: explain why!

\subsection{LM as Generator: Multi-attribute Prompt}
Our final prompting technique focuses on retrieving information about a set of attributes, by querying multiple attributes together. This technique combines strengths of the zero-prompt and the single-attribute prompt models, as it is able to both verbalize the attributes and capture correlations across them. Unlike other mechanisms, this method allows us to control the information content per instance, by varying the set of queried attributes. As we show in sections \ref{experiments} and \ref{k_prompt}, varying the attribute queries across training instances is crucial to achieve generalization. %Finally, our few-shot experiments suggest that models trained with this technique have the ability to extract reasoning mechanisms instead of learning attribute-specific information. 

For this technique, the prompt lists the attributes that the model should consider. This list is dataset specific and can vary between training and testing (i.e., out-of-domain) or even across training instances. The model is trained to generate the attributes that changed, as shown in Figure \ref{fig:prompts}. This technique works with text generation models and was tested on both T5 (fine-tuned) and GPT-3 (few-shot).

The first version of this model, the \emph{all-attribute prompt}, queries all attributes that could change in the same instance. However, the risk with this approach is that, because the prompt is fixed, the model learns to pay little attention to the specific attributes that appear in it. We therefore propose a variant of this method, the \emph{$k$-attribute prompt}, aiming to achieve high performance in both in-domain and out-of-domain scenarios. The objective is to learn about attribute dependencies but also force the model to pay attention to the specific attributes being prompted. To achieve this, we prompt the model with $k$ random attributes and train it to predict changes \emph{only} among these $k$ attributes. More specifically, for each training example, we partition the 51 attributes into $q$ random groups where $q$ is a random integer between 1 and 5. $k$ refers to the number of attributes in each partition. This method ensures that the model is queried with $k$ random attributes and that all 51 attributes are always queried for each example.
%We note that only a few examples were enough to ensure that the model correctly learns the following format

%This prompting method lets us evaluate some characteristics of the reasoning patters captured by language models. Our experiments show the importance of prompt content as a means of sharing task information with the model. We explore how we can use this technique to achieve   

%This prompting method lets us evaluate some characteristics of the reasoning patters captured by language models. First, the model is given both verbalizations of attributes and correlations across attributes, combining the strengths of both the zero-prompt and single-prompt methods. Second, via the few-shot model we test the ability to extract a reasoning mechanism instead of learning attribute-specific information. This is because the model needs to generate the changed attributes, given only a small list of examples that may contain different attributes. Finally, this technique could also improve the generalization abilities of a fine-tuned model in unseen attributes, by changing the prompt that communicates the verbalization of the attributes to the $k$-attribute model.     

\begin{table*}[ht]
\centering
\begin{tabular}{@{}l@{\hspace{2em}}ccc@{\hspace{3em}}ccccc@{}}

 & \multicolumn{3}{c}{All attributes}                                         & \multicolumn{5}{c}{Per-attribute F1}                                                                                               \\ 
\multicolumn{1}{c}{Model}& Pr & Re & F1 & Dist & Size & Mass & Temp & isBroken \\ 
 \toprule
 Physical Interaction, (PiGLET)   &   97.4   &  91.6     & \textbf{94.4}   & \textbf{93.6}  & 79.2  & 98.3  & \textbf{99.6}  & 92.8 \\ 

\midrule
n-gram LogReg (baseline)                   & 87.8 & 88.0  & 87.9   & 78.8  & 74.7  & 97.8  & 94.0  & 79.4 \\
RoBERTa-base, zero-prompt        & 95.2 & 92.6  & 93.9   & 90.6  & 82.7  & \textbf{100.0} & 95.3  & \textbf{94.7} \\ 
T5-base, all-attribute prompt  &   93.0   &   95.4    &   94.1     &   91.7    &   \textbf{83.5}    &     \textbf{100.0}  &    95.8   &  90.3     \\ 

\bottomrule

\end{tabular}
\caption{Micro-Precision, Recall and F1 scores across all 14 attributes in PiGLET. Per-attribute F1 scores for challenging attributes, as in \cite{zellers-etal-2021-piglet}. Language-only models perform competitively with PiGLET.}
\label{tab:piglet}
\end{table*}

%\textcolor{blue}{would prefer piglet results on first page and open pi on the next}
\section{Experiments \& Results}
\label{experiments}

%\subsection{Metrics}

%Our task is a multi-label classification where, given some context and an entity of interest, we need to identify which attributes change. For most pairs \textit{context, entity}, event implications affect only 1-2 attributes. This results in a few positive instances (i.e., attributes that change) and a large number of negative instances (i.e., attributes that do not change). Furthermore, we observe that the number of positive instances significantly varies across attributes: for example, in the training set of Open PI, \textit{location} has 4505 positive instances, while \textit{distance} only 53.

Our task is a multi-label classification where, given some context and an entity of interest, we need to identify which attributes change. Due to the significant label imbalance, in our experiments we report micro- Precision, Recall, and F1 for the positive instances, across labels. In addition to these metrics, we measure per-attribute Precision, Recall and F1 for both datasets (details in \autoref{sec:attr_performance}).

\subsection{PiGLET}
\paragraph{Baselines:} The strongest baseline is the PiGLET model, which is a combination of physical interaction and language model, based on GPT-2 \cite{radford2019language}. It was proposed in the paper introducing the dataset and is currently state-of-the-art. Unlike the other models, it learns by interacting with a simulator and has access to the pre-state of each entity. We also use a simple n-gram Logistic Regression baseline to both establish the overall difficulty of the dataset and measure benefits due to the pre-training of LLMs.

%Second, we use two of our proposed models: a classifier (zero-prompt RoBERTa) and a generative model (multi-attribute T5). These two models help us to test the following hypothesis: (1) whether a model can learn only from language data and (2) how task formulation influences the performance of a LLM. Finally, we use a relatively naive baseline (n-gram Logistic Regression) to determine the performance gain due to the pre-training of the language model.     

\paragraph{Results:} As shown in \autoref{tab:piglet}, all models perform relatively well on the PiGLET dataset. The extremely small margin in performance between Physical Interaction and the proposed models (RoBERTa zero-prompt and T5 all-attribute) indicates that language models can learn about physical attributes even without the need of physically interacting with the environment. However, we should highlight that this conclusion holds for datasets similar to PiGLET and the importance of physical interactions remains an open question that must be tested in more realistic and challenging datasets.

Despite the high performance of our proposed models, previously reported baselines on PiGLET show significantly lower performance than the Physical Interaction model. Notably, their baseline using T5-base achieves only 53.9\% in hard accuracy, compared to 81.1\% of the Physical Interaction model \cite{zellers-etal-2021-piglet}. Unfortunately we cannot directly compare these results to our proposed models due to their choice of metric (hard accuracy) and different problem formulation, where the input and output is the encoding of the pre- and post-state of the entity. 
Despite the use of different metrics, we observe a minimal performance difference between language-only models and PiGLET. This highlights the importance of using proper prompting techniques and task formulation to take full advantage of LLMs and draw valid conclusions. 
% However, the fact that this large gap in performance disappears in our setup highlights the importance of using proper prompting techniques and task formulation that enhance the abilities of a LLM.% This proves our initial claim that researchers must use baselines in a way that enhance a model's capabilities in order to ensure a fair comparison across models.

Our final observation is that there is a larger gap between the n-gram LogReg model and the rest of the models. This shows that, although language is very useful to predict physical event implications, pre-trained language models still have an advantage due to the information they have previously seen. This raises the question of how can we better exploit the relations that pre-trained language models already know, which we explore via the next set of experiments.

\subsection{OpenPI}

% Please add the following required packages to your document preamble:
% \usepackage{multirow}
\begin{table*}
\centering
\begin{tabular}{llccc@{\hspace{2em}}ccc}
                            &                                        & \multicolumn{3}{c}{In-domain}                                & \multicolumn{3}{c}{Out-domain}             \\ 
Training                    & Model                                  & Pr         & Re         & F1         & Pr & Re & F1 \\ 
    \toprule                            
% \multirow{4}{*}{Fine-tuned} & \begin{tabular}[c]{@{}l@{}}RoBERTa-base,\\ zero prompt\end{tabular}             & 54.8}       & 47.4}       & 50.8       & -}   & -}   &   - \\ 
Zero-shot                   & RoBERTa-large, single-attribute prompt & 3.1  & 63.3  & 5.9  & 2.4  & 68.8  &  4.6  \\ 
\midrule

\multirow{2}{*}{Few-shot}   & GPT-3-Babbage, single-attribute prompt & 3.7  & 82.4  & 7.1  & -    & -     & -    \\ 
                            & GPT-3-DaVinci, all-attribute prompt  & 37.6 & 24.5  & 29.7 & 28.3 & 12.9  & 17.7 \\ 
\midrule
\multirow{4}{*}{Fine-tuned} & GPT-2 (baseline in Open PI) & 49.8 & 11.8 & 19.1 & - & - & - \\

& RoBERTa-large, zero prompt             & 65.1 & 40.1  & 49.6 & -    & -     & -   \\ 
                            & RoBERTa-base, single-attribute prompt  & 40.3 & 55.1  & 46.6 & 21.3 & 26.2  & \textbf{23.5}  \\ 
                            & T5-base, single-attribute prompt       & 34.6 & 53.3  & 42.0 & 15.9 & 21.5  & 18.2   \\ 
                            & T5-base, all-attribute prompt        & 47.5 & 56.0  & \textbf{51.4} & 25.0  & 1.2  & 2.2   \\ 
                            & T5-base, $k$-attribute prompt & 52.8 & 50.0 & \textbf{51.4} & 16.8 & 22.7 & 19.3\\
\bottomrule

\end{tabular}
\caption{Micro-Precision, Recall and F1 scores for Open PI. In-domain attributes refers to the 51 originally curated attributes, while out-domain to the 41 attributes introduced by human annotators.}
\label{tab:openpi}
\end{table*}

\label{openpi_exp}

Since our results in PiGLET show that it is not a challenging dataset, we use Open PI to compare the proposed prompting techniques. With the exception of the GPT-3 models, all models have relatively similar sizes, ranging from 123M (RoBERTa-base) to 354M (RoBERTa-large) parameters.  
%This distinction helps us to draw conclusions about the effect of pre-training and determine what a model already knows about physical attributes.

%With the exception of the GPT-3 models, all our models have relatively similar sizes, ranging from 123M(RoBERTa-base)-354M(RoBERTa-large) parameters. This ensures that we have as fair comparison of the models and the prompting mechanism as possible. 

\textbf{Few-shot:} For each instance in the test set, we pick 10 examples from the training set to be included in the prompt - there are marginal improvements beyond four \citep{min2022rethinking}. Performance in complex tasks like QA is sensitive to prompt selection \citep{liu-etal-2022-makes}. Following previous work, we pick the relevant examples based on semantic similarity \citep{reimers-gurevych-2019-sentence}. In the single-attribute prompt setting, we include examples querying the same attribute, and balance both positives and negatives.

% \citet{min2022rethinking} discussed the optimal number of prompts for few shot learning. Generally the more examples are included the better the model performs. However, when more than four are included the benefits are marginal. \citet{liu-etal-2022-makes} experiments with different ways to choose prompts for GPT-3. While in sentiment analysis the choice of prompts was not of significant importance, in QA the benefits were greater. They recommend using semantic similarity to find the most relevant examples to use as prompts.

\textbf{In-domain vs out-domain:} All our models are trained on the initial 51 attributes (\autoref{data:openpi}). For in-domain experiments, the models are tested on the same set of attributes, while for out-of-domain on the new attributes introduced by human annotators. After removal of rare attributes and merging of synonyms, the out-of-domain set consists of 41 unique attributes.

\paragraph{Results:} As shown in Table \ref{tab:openpi}, the best performing models in-domain are the multi-attribute prompt models. The performance difference between the multi-attribute models and the zero-prompt baseline shows that the verbalization of attributes has a positive impact on performance, which is further supported by our findings in \autoref{sec:frequent}. Furthermore, our models beat the GPT-2 model, proposed by \citet{tandon-etal-2020-dataset} along with the Open PI dataset. This model generates sentences describing entity state changes but, unlike our models, does not verbalize the attributes. Finally, we observe a drop in performance for both T5 and RoBERTa single-attribute prompt, which confirms that attribute dependencies are important in our task.  

%We hypothesize that multi-attribute models perform best because they see multiple attributes in the same instance and, thus, are able to learn correlations across them. On the other hand, the single-attribute models only query one attribute at a time, resulting in the loss of such information. To verify that model size differences do not impact our results, we also did experiments with RoBERTa-base zero-prompt, which shows very similar performance to RoBERTa-large zero-prompt.

Despite its good performance in previously seen attributes, the zero-prompt model cannot classify out-of-domain attributes because its output is a fixed-dimension binary vector. The best out-of-domain performance is achieved by RoBERTa single-attribute, followed by T5 $k$-attribute prompt.

We observe that, despite the very low out-of-domain performance of the T5 all-attribute prompt model, the other two variants of the same prompting technique (GPT-3 all-attribute and T5 $k$-attribute) perform competitively. This confirms our hypothesis that fine-tuning with a fixed query hurts the generalization properties of the model, something that can be avoided with few-shot learning or by shifting focus to different attributes during training (i.e., single-attribute or $k$-attribute). 

%The T5 $k$-attribute prompt model is further discussed in Section \ref{k_prompt}, as the best performing model overall with respect to both in-domain and out-domain experiments.  

\section{Discussion}
We further study the models' behavior with respect to the type of attributes they see and their generalization properties. This analysis serves to uncover advantages and disadvantages of each technique and suggest promising methods for future work to enhance both model performance and robustness. 

For all our experiments we use Open PI. Due to its greater diversity of attributes and larger size, it is a better candidate than PiGLET to analyze the limitations of the models.

% \begin{figure*}[!ht]
%      \centering
%      \begin{subfigure}[b]{0.32\textwidth}
%          \centering
%          \includegraphics[width=\textwidth]{EMNLP 2022/figures/zero-prompt_attr-frequencies.pdf}
%          \caption{Zero-prompt RoBERTa.\\
%          Pearson r= 0.74}
%          \label{fig:zero_prompt_freq}
%      \end{subfigure}
%      \hfill
%      \begin{subfigure}[b]{0.32\textwidth}
%          \centering
%          \includegraphics[width=\textwidth]{EMNLP 2022/figures/single-attr-roberta_attr-frequencies.pdf}
%          \caption{Single-attribute prompt RoBERTa.\\
%          Pearson r = 0.65}
%          \label{fig:single_attr_freq}
%      \end{subfigure}
%      \hfill
%      \begin{subfigure}[b]{0.32\textwidth}
%          \centering
%          \includegraphics[width=\textwidth]{EMNLP 2022/figures/multi-attr-t5_attr-frequencies.pdf}
%          \caption{Multi-attribute prompt T5.\\
%          Pearson r= 0.56}
%          \label{fig:multi_attr_freq}
%      \end{subfigure}
%         \caption{F1 score per attribute, as a function of the number of positive instances in the training data. We estimate the Pearson correlation between the frequency of an attribute and its F1 score.}
%         \label{fig:freq_graphs}
% \end{figure*}

\subsection{Reasoning with Rare Attributes}
\label{sec:frequent}
Since some attributes are significantly more frequent than others, fine-tuned models have been exposed to more data about them, which influences performance. For example, performance across all fine-tuned models for the most frequent attribute \textit{location} is substantially higher compared to other attributes (F1 = 0.65-0.75). Although most models are expected to perform well on such high frequency attributes, our analysis provides useful insights on the models' ability to learn reasoning patterns in limited-data scenarios. 

\begin{figure}
    \centering
    %trim={<left> <lower> <right> <upper>}
    % \includegraphics[width=0.5\textwidth,trim={0 11.5cm 0 0},clip]{EMNLP 2022/figures/attribute_freq_analysis.pdf}
    \includegraphics[width=\columnwidth,trim={0 13.7cm 0 0},clip]{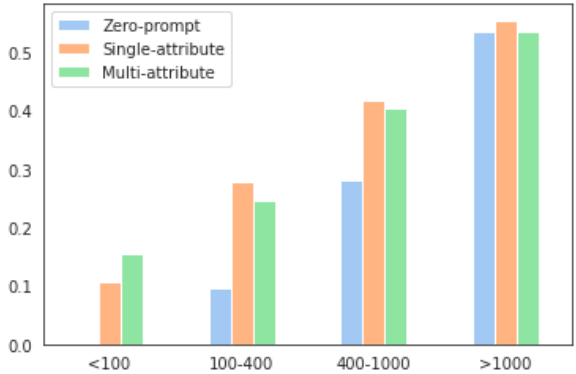}
    \caption{Performance per attribute frequency in training data. Each bar shows the weighted-F1 score across all attributes in the same frequency category.}
    \label{fig:freq_graphs}
\end{figure}

We study per-attribute model performance based on each attribute's frequency in training data for the three prompting techniques: RoBERTa zero-prompt, RoBERTa single-attribute, and T5 all-attribute. After clustering each attribute with respect to its frequency and its F1 score, we observe four distinct clusters: low (<100 instances), medium-low (100-400), medium-high (400-1000) and high (>1000) frequency. In Figure \ref{fig:freq_graphs} we plot the weighted-F1 score per cluster for the three models. Our first observation is that performance across all models increases for attributes with higher frequency. This conclusion is also supported by the per-attribute Spearman correlation between performance and frequency, shown in Table \ref{tab:correlation}. This confirms our hypothesis from PiGLET that LLMs can learn physical interactions and achieve higher performance when there is sufficient labeled data to fine-tune on.

\begin{table}
\resizebox{\columnwidth}{!}{
\begin{tabular}{lccc}
\hline 
 & \begin{tabular}[c]{@{}c@{}}RoBERTa, \\ zero-prompt\end{tabular} & \begin{tabular}[c]{@{}c@{}}RoBERTa, \\ single-attribute\end{tabular} & \begin{tabular}[c]{@{}c@{}}T5, \\ all-attribute\end{tabular} \\ \hline
\begin{tabular}[c]{@{}l@{}}Spearman \\ correlation\end{tabular} & $\rho$ = 0.82                                                   & $\rho$ = 0.80                                                        & $\rho$ = 0.51                                                  \\ \hline
\end{tabular}}

\caption{Spearman correlation between attribute frequency and F1 score. High correlation means the model learns primarily high-frequency attributes. All results have p-value < 0.001.}
\label{tab:correlation}
\end{table}

Our second observation is that, although performance in high-frequency attributes is similar across all models, it significantly drops for RoBERTa zero-prompt when frequency decreases. This shows that the model struggles to learn with fewer examples. This difference is most striking in the low-frequency cluster, where the model learns nothing (F1 = 0.0). On the other hand, both RoBERTa single-attribute and T5 all-attribute have relatively high performance in low-frequency attributes, where some attributes are easier to learn than others. This supports one of our main hypothesis in this paper that, by \emph{verbalizing and querying specific attributes, models pay attention to each attribute and learn reasoning patterns, a crucial step in limited-data scenarios}.

\subsection{Prompt Diversification via the $k$-attribute Prompt Model}
\label{k_prompt}
\begin{figure}
    \centering
    \includegraphics[width=\columnwidth,trim={0 0.5cm 0 0},clip]{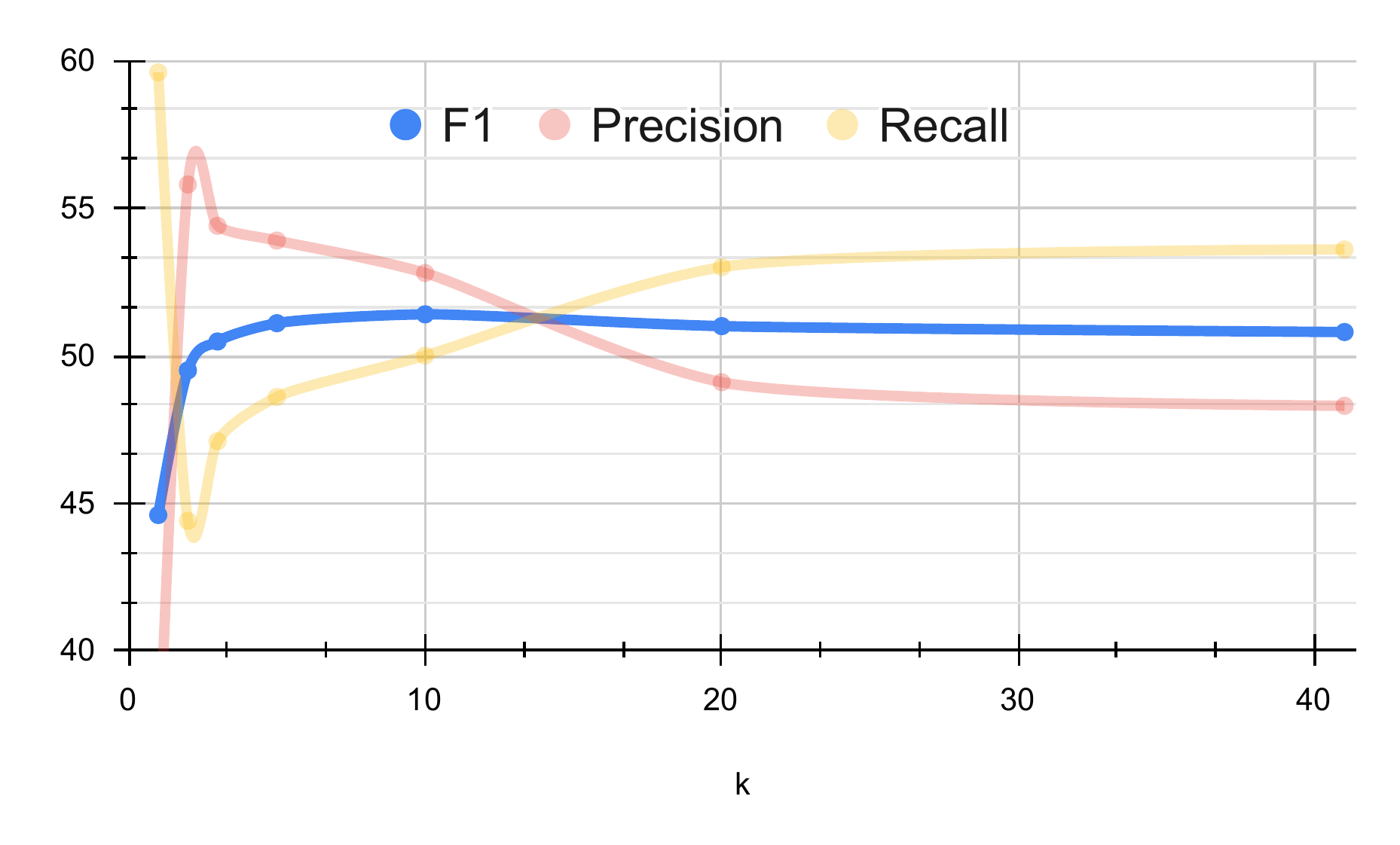}
    \caption{F1, Precision, and Recall scores as a function of the number of attributes used in the prompt during evaluation for the $k$-attribute model}
    \label{fig:k-attribute}
\end{figure}

Through manual inspection we find that the all-attribute models have an inherent bias towards generating attributes that appeared in the training data, even when prompted with new ones. Their performance is in fact poor in the out-of-domain setting (2.2 F1, \autoref{tab:openpi}). Now the question is whether this is a limitation of the reasoning abilities of the multi-attribute models or a bias introduced by its training scheme.

We propose the $k$-attribute model to alleviate training biases by randomizing the queried attributes. Notably, this model still maintains the core assumptions behind the multi-attribute prompt model of querying multiple attributes at once. We observe that this simple technique results in the same in-domain F1 score as the all-attribute prompt model, while significantly improving its out-of-domain performance. 
This shows that the observed limitations with the all-attribute prompt model are due to training biases that prevent the model from generalizing to unseen attributes.

Once trained, the $k$-attribute prompt model can be queried with varying number of attributes. In \autoref{fig:k-attribute}, we plot the performance of the model as a function of the number of attributes used in the query during evaluation. We observe a drop in performance when the model is queried with a single attribute (similar to the single-attribute prompt models). The performance is highest around 10 attributes and drops slightly beyond that. We also observe that by varying $k$, we can modulate precision and recall, suggesting that there are both lower and upper bounds on the optimal number of attributes that LLMs can consider at once.

We also experimented by grouping attributes in a prompt based on their semantic similarity, but this did not yield any significant changes in performance. We leave it to future work to investigate further how to optimally choose the groups to use in a prompt during training and inference.

\subsection{Semantic Similarity and Generalization}
\label{sec:sem_similarity}

A major obstacle for NLP models is to apply the reasoning patterns they have learned to unseen attributes. Although the overall performance is lower in out-of-domain (best F1 = 23.5) compared to in-domain experiments (best F1 = 51.4), we observe that it varies significantly across different attributes. In this part of our analysis, we investigate the models' generalization abilities to out-of-domain attributes, based on their relation to in-domain attributes. 

Essentially we identify two types of out-of-domain attributes: (1) these that are semantically similar to some in-domain attribute(s), and (2) these that have no similarity to any in-domain attribute. These two groups of attributes also evaluate the degree of the model's generalization abilities, as it is easier to generalize to different verbalizations of a previously seen attribute than to a completely new concept. For this part of the analysis we use the RoBERTa single-attribute prompt model, as it has the best out-of-domain performance. 

To identify related attributes, we firstly use cosine similarity distance on top of an encoder trained for semantic similarity \citep{reimers-gurevych-2019-sentence}. After manual curation, we identify 21 out-of-domain attributes that are closely related to in-domain attributes (Group Matched), as we see in \autoref{tab:synonyms}. The 20 remaining out-of-domain attributes are more dissimilar and do not have matching in-domain attributes (Group Dissimilar).

For each of the two groups (Group Matched and Group Dissimilar), we estimate the weighted-F1 score. We observe that Group Matched reaches \textbf{F1 = 29.4}, while Group Dissimilar \textbf{F1 = 13.6}. For Group Matched, we also verify that the model's performance on closely related attributes is similar by measuring their Pearson correlation, which is $r=0.67$ ($p$-value $<0.05$). Both results indicate that \textit{the model understands the semantics of the attributes despite different verbalizations, however, it struggles with more complex reasoning mechanisms, such as applying the acquired patterns to entirely new attributes}.

\subsection{Challenging Semantic Types}

\begin{figure}
     \centering
    \includegraphics[width=\columnwidth]{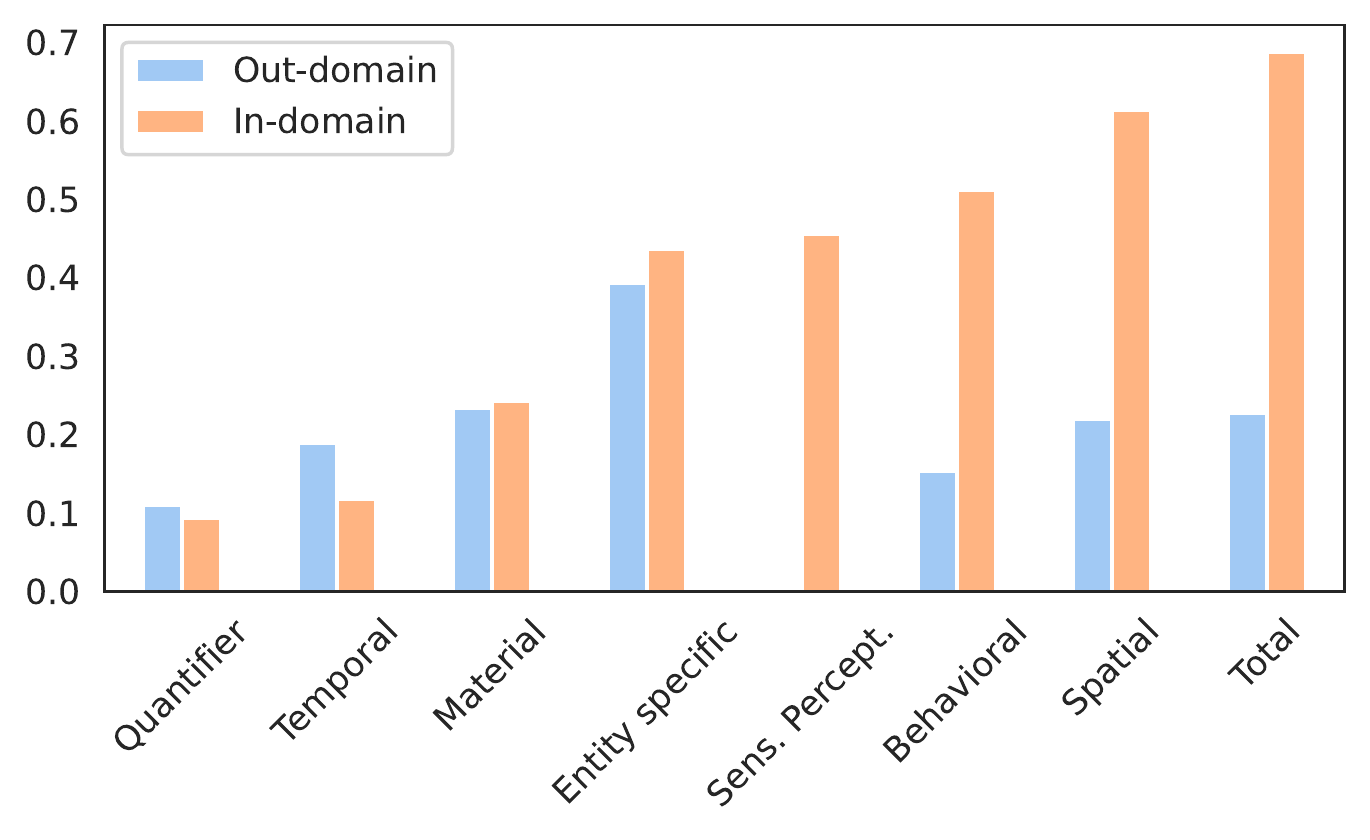}
    \caption{F1 scores per attribute semantic type.}
    \label{fig:sem_clusters}
\end{figure}

In this part of our analysis, we explore why some classes of physical attributes appear to be inherently more difficult for LLMs. More specifically, we manually design an ontology of attributes into seven major semantic types and then group each in-domain and out-of-domain attribute according to the information it encodes, as seen in \autoref{app:sem_clusters}. Via this analysis we aim to identify evaluate each semantic type with respect to: (1) in-domain performance, and (2) generalization to unseen attributes. For this analysis we use RoBERTa single-attribute prompt, as it has the best out-of-domain performance.

\autoref{fig:sem_clusters} shows that the model particularly struggles to predict attributes of the \emph{Quantifiers} and \emph{Temporal} semantic types (in-domain). These attributes are known to be challenging for current LLMs \citep{ravichander-etal-2019-equate}.

We further observe that the \emph{Entity-specific} and \emph{Material} semantic types are equally challenging for both in-domain and out-of-domain attributes. These semantic types describe inherent properties of an entity, such as \textit{fullness}, that can only change due to very specific events, such as \textit{put X into Y}. On the other hand, the \emph{Spatial} and \emph{Behavioral} types show a large discrepancy between in-domain and out-of-domain performance. This is surprising given that these semantic types contain high-frequency attributes, like \textit{location}. This highlights \textit{the limitations of current models to predict physical changes outside of controlled environments}.

% PREVIOUS PAPER

%\paragraph{Challenging semantic types} 
%Second, we study performance on different classes (semantic types) of attributes. We distinguish seven major semantic types based on the information each attribute encodes (\autoref{app:sem_clusters}). 

%As shown in Figure \ref{fig:sem_clusters}, some semantic types are equally challenging for both in-domain and out-of-domain attributes. These semantic types tend to describe inherent properties of an entity that can be changed by very specific events, such as \textit{fullness}. Notably, we also show that the model struggles to capture \emph{Quantifiers} and \emph{Temporal} semantic types, which are known to be challenging for current LLMs \citep{ravichander-etal-2019-equate}. 

%On the contrary, other semantic types have a large discrepancy between in-domain and out-of-domain performance, despite in some cases, having the most training instances, like \emph{Spatial} attributes. With \emph{Behavior}, which relates to the behavior of an entity during an event (e.g., speed), both semantic types contain attributes that could be affected by many event types. This suggests that the performance at predicting changes-of-state for a given attribute \emph{relates to the diversity of event types that could affect the attribute}. Finally, our analysis highlights the limitations of current models to \emph{predict physical changes outside controlled environments} and \emph{apply what they learned in out-of-domain settings}.

\subsection{Error Analysis}

To identify the cause of low out-of-domain performance and study the models' generalization abilities, we perform a manual error analysis of out-of-domain outputs from the best performing models: T5 $k$-attribute and RoBERTa single-attribute prompt.

We identify four major types of errors indicating a varying degree of understanding of context and entities involved. The results of this analysis are shown in \autoref{tab:error_analysis}.\\[-5pt]

\noindent \textbf{False negatives}: correct predictions that are missing from the annotations. This error type does not reflect a failure of the models, but rather of the dataset which was crowd sourced. Since out-of-domain attributes were introduced by workers on Amazon Mechanical Turk, each annotator may introduce attributes that were not considered by others while annotating different instances. This is particularly prominent among similar concepts, such as \textit{width} and \textit{size}, which oftentimes change together. As we see in \autoref{tab:error_analysis}, \emph{false negatives} are responsible for 41.5\% of errors made by T5 k-attribute prompt and 25.4\% of those made by RoBERTa single-attribute. This highlights that the gap between out-of-domain and in-domain performance is narrower than what our automated evaluation showed.

\textit{False negative} errors can be divided into two subcategories. The first category accounts for predicted attributes that are synonyms of the annotated attributes and could replace them in the particular instance. The second category comprises predicted attributes that significantly differ but complement the annotated attributes, such as \textit{flexibility} and \textit{size}. We found that the first category of synonyms is responsible for 53\% (T5 k-attribute prompt) and 44\% (RoBERTa single-attribute) of the instances with \textit{false negative} errors. \\[-5pt]
%This results in attributes that frequently change together, but the annotators failed to identify both in the same instance. We found that almost 53\% (T5 k-attribute prompt) and 44\% (RoBERTa single-attribute) of the instances in this error category refer to such frequently together occurring attributes.

\noindent \textbf{Wrong context}: predictions that could be correct for the given entity, but incorrect given the context. This error represents the models' challenges with respect to event implications and reasoning.\\[-5pt]

\begin{table}[ht]
\centering
\resizebox{\columnwidth}{!}{
\begin{tabular}{lcc}
\hline
\textbf{Error Type} & \textbf{T5} & \textbf{RoBERTa}\\
 & \textbf{k-attribute} & \textbf{single-attribute}\\
\hline
False negatives & 41.5\% & 25.4\% \\
Wrong context & 7.6 \% & 6.5\%\\
Wrong entity  & 2.7\% & 20.7\%\\
No prediction & 48.2\% & 47.4 \% \\
 \hline
\end{tabular}
}
\caption{Error categories and prevalence of each category as a percentage of the number of instances. Based on out-of-domain attributes. \textit{Wrong context} implies the prediction could be correct for the given entity but is incorrect in the given context. \textit{Wrong entity} means the attribute change does not apply to the given entity in any context.}
\label{tab:error_analysis}
\end{table}

\noindent \textbf{Wrong entity}: wrong attribute change predictions for the given entity in any context. This is the most severe error since it shows that the model is not able to link the attributes to the entity. While this error is very rare for the T5 k-attribute model (only 2.7\%), it is frequent for the RoBERTa single-attribute model (20.7\%).\\[-5pt]

\noindent \textbf{No prediction}: instances with null predictions. This is the most frequent error type for both models, accounting for almost half of the errors. This error occurs when the model decides that there is no attribute change from the given list of attributes, which results in a significant drop in recall. This highlights that both models struggle to identify which out-of-domain attributes are relevant to a particular context and entity. \\

\section{Conclusion}
Predicting physical changes due to events is a challenging problem for current models, especially in out-of-domain or limited-data scenarios. We show that, by using proper task formulation, LLMs can learn physical event implications even without physical interactions. Future work should explore the question of whether physical interactions are necessary in more complex and realistic settings, by (1) providing more challenging datasets that test the model limitations, and (2) ensure a fair comparison of the language-only baselines. 

Furthermore, we show that the performance of a LLM may significantly vary based on how we use it, and, overall, LLMs can benefit from: (1) verbalizing the attributes, (2) varying the prompt information content across instances, and (3) querying multiple attributes in the same instance. By following these guidelines, we show significant improvements in unseen attributes and attributes of low-frequency. Last, our error analysis and discussion sections provide useful insights for future work, with respect to prompt content and shortcomings of the current datasets that study physical event implications.

%Predicting physical changes due to events is a challenging problem for current models, especially in out-of-domain or limited-data scenarios. We showed that, by using proper task formulation, LLMs can learn physical event implications even without physical interactions. Future work should explore the question of whether physical interactions are necessary in more complex and realistic settings. We show that LLMs can benefit from: (1) verbalizing the attributes, (2) varying the prompt information content across instances, and (3) querying multiple attributes in the same instance. By following these guidelines, we show significant improvements in out-of-domain and limited-data scenarios (i.e., rare attributes). Lastly, our analysis of semantically similar attributes can provide useful insights for future work, both with respect to prompt content and new datasets that study physical event implications.     

\section{Acknowledgments}

We thank the anonymous reviewers for their helpful feedback. We also thank Alan Ritter and Lori Levin for their comments and feedback.  

\section{Limitations}

\paragraph{Computing resources} 
The different prompting methods have trade-offs in terms of computational costs. In particular, the all-attribute and zero-attribute query all changes at once. With the $k$-attribute prompt, we query attributes in smaller groups requiring on average \#attributes$/k$ times more computations than for the all-attribute model (in our case five times). The single-attribute model encodes each attribute separately requiring \#attributes times more computations. We were unable to test GPT-3 for single-attribute because of the cost of the larger number of queries it would have required. The experiments that did not involve GPT-3 were run on two NVIDIA K-80 GPUs with 12Gb memory. 
% Computational resources require for the single-attribute models are greater compared to the multi and $k$ attribute models.

\paragraph{Dataset limitations} Given the complex nature the event implication task, both datasets have several limitations. PiGLET, which is based on a virtual environment, has relatively simple language that is not representative of naturally occurring text. Furthermore, because it is a relatively small dataset with respect to number of attributes and entities, the training set covers a large subset of the possible configurations in that virtual environment. This explains the very high performance of all models.

Although Open PI does not suffer from such limitations, we discovered several inconsistencies in the annotations. These inconsistencies mainly involve: (1) wrong attributes, (2) inconsistent labeling, and (3) duplication of attributes. Although we manually edited several of these problems by merging and filtering attributes, we could not address the inconsistencies in labeling. This resulted could have influenced model performance.

\paragraph{Automatic Prompt Generation}
In this work, we did not explore whether prompts can be automatically generated. There have been several recent studies aiming at generating either discrete or soft prompts \citep{shin-etal-2020-autoprompt, lester-etal-2021-power}. In our case, the changes in information content involved a deeper understanding of the task and required human involvement. As the field of prompt generation matures, future work could investigate automating the process of finding prompts with variable information content.

\paragraph{Multi-task learning}
We do not directly explore benefits from multi-task learning even though \citet{raffel2020exploring, wei2021finetuned} show that this can significantly improve zero-shot and few-shot performance. However, the GPT-3 model that we used in our experiments is the Instruct GPT-3 model which is the result of additional prompt-based finetuning.

\bibliography{anthology,custom}
\bibliographystyle{acl_natbib}

\newpage
\appendix

\section{Appendix}
\label{sec:appendix}
Our experiments are built on top of the Huggingface library \citep{wolf-etal-2020-transformers}.

\subsection{Metrics}
Our task is a multi-label classification where, given some context and an entity of interest, we need to identify which attributes change. For most pairs \textit{context, entity}, event implications affect only 1-2 attributes. This results in a few positive instances (i.e., attributes that change) and a large number of negative instances (i.e., attributes that do not change). Furthermore, we observe that the number of positive instances significantly varies across attributes: for example, in the training set of Open PI, \textit{location} has 4505 positive instances, while \textit{distance} only 53.
Due to the significant label imbalance, in our experiments we report micro- Precision, Recall, and F1 for the positive instances, across labels. In addition to these metrics, we measure per-attribute Precision, Recall and F1 for both datasets. 

\subsection{Hyperparameters}

We performed hyperparameter search in the following way. Based on the model size, we picked the largest batch size that could fit on our GPUs. Then we performed hyperparameter search on the dev set (6 values in range $[10^{-3}, 10^{-6}]$), label smoothing (0, 0.1, 0.2) via grid search. We report in \autoref{tab:hyperparameters} the hyperparameters we use in each case. We used the default values in the transformer library for the rest. For T5 we also varied the task prefix and its position based on the relevant pre-training tasks, without observing significant differences. We use Adam with betas (0.9,0.999) and $\epsilon=$1e-08 for T5 experiments. The runtime for each hyperparameter combination in Open PI is: about 2 hours for multi-attribute, about one hour for zero-prompt, about two days for single-attribute (T5 and RoBERTa have similar runtime).

\begin{table}[H]
    \centering
    \resizebox{\columnwidth}{!}{
    \begin{tabular}{ll c c c c}
    \toprule
          Data & Model & Epochs & Batch size & Learning Rate & Label Smoothing \\
         \midrule
         PiGLET & RoBERTa, &  &  & & \\
         & zero-prompt & 30 & 20 & 4e-05 & 0.0 \\
         & T5 all-attr   & 50 & 32 & 3e-05 & 0.1\\
         \hline
         Open PI &RoBERTa, &  &  & & \\
         & zero-prompt & 20& 32 & 1e-05 & 0.0\\
         & RoBERTa, &  &  & & \\
         & single-attr & 6& 16 & 1e-05 & 0.1\\
         & T5 single-attr  & 8 & 16 & 5e-05 & 0.1\\
         & T5 all-attr   & 8 & 16 & 5e-05 & 0.1 \\
         & T5 $k$-attr   & 10 & 16 & 5e-05 & 0.1 \\

         \bottomrule
    \end{tabular}}
    \caption{Hyperparameters     }
    \label{tab:hyperparameters}
\end{table}

To verify that model size differences do not impact our results, we also did experiments with RoBERTa-base zero-prompt, which shows very similar performance to RoBERTa-large zero-prompt.

\subsection{In-domain Attributes and their Frequency}

\begin{table}[H]
    \centering
    \resizebox{0.85\columnwidth}{!}{
    \begin{tabular}{lrrr}
\toprule
                 Attribute &  Train &  Dev &  Test \\
\midrule
             location &   4505 &  360 &   803 \\
            cleanness &   1255 &  117 &   167 \\
              wetness &   1211 &   80 &   215 \\
          temperature &   1184 &   91 &   184 \\
               weight &   1073 &   84 &   124 \\
             fullness &    694 &   62 &   122 \\
               volume &    676 &   56 &   174 \\
          composition &    662 &   48 &    90 \\
                shape &    538 &   55 &    65 \\
              texture &    515 &   34 &    74 \\
            knowledge &    409 &   27 &   119 \\
          orientation &    330 &   15 &    45 \\
                color &    292 &   13 &    33 \\
                 size &    264 &   26 &    50 \\
                power &    245 &   11 &    18 \\
         organization &    242 &   14 &    37 \\
               motion &    242 &   15 &    33 \\
            ownership &    212 &    6 &    19 \\
         availability &    195 &   30 &    63 \\
                 step &    171 &    8 &    13 \\
                speed &    151 &    3 &    18 \\
             pressure &    148 &    4 &    14 \\
                taste &    145 &    8 &    14 \\
               length &    122 &    9 &    17 \\
electric conductivity &    121 &    9 &    18 \\
                smell &    120 &    7 &    43 \\
                sound &     68 &    6 &     6 \\
           brightness &     65 &    0 &     7 \\
            thickness &     64 &    4 &    16 \\
             strength &     64 &    2 &    14 \\
             hardness &     63 &    5 &    10 \\
                skill &     62 &    3 &     4 \\
             openness &     55 &    2 &    16 \\
             coverage &     54 &    3 &     7 \\
            stability &     54 &    6 &    14 \\
                focus &     53 &    4 &     5 \\
                 cost &     53 &    6 &     9 \\
             distance &     53 &    0 &    11 \\
           appearance &     44 &    8 &     8 \\
           complexity &     44 &    1 &     5 \\
               amount &     40 &    3 &    16 \\
\bottomrule
\end{tabular}}
    \caption{Attribute occurrences in training, validation, and test sets.}
    \label{tab:my_label}
\end{table}

\subsection{In-domain performance, per-attribute}

\label{sec:attr_performance}
\begin{figure}[ht]
    \centering
    \includegraphics[width=\columnwidth]{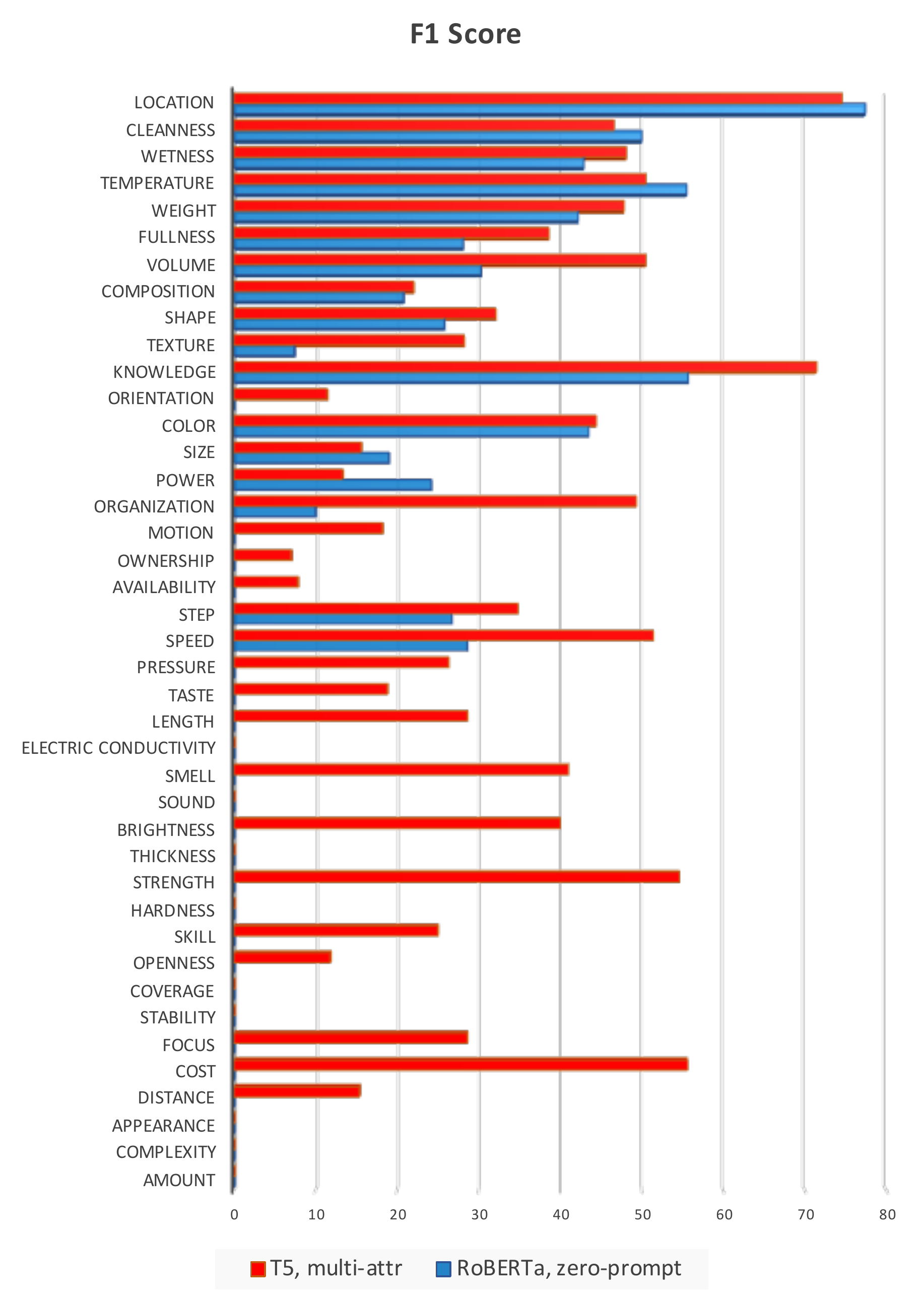}
    \caption{F1 score per attribute for RoBERTa zero-prompt and T5 multi-attribute prompt models in Open PI.}
    \label{fig:attr_f1_openpi}
\end{figure}

In \autoref{fig:attr_f1_openpi} we show the in-domain F1 score per attribute for RoBERTa zero-prompt and T5 multi-attribute prompt models in Open PI. The attributes are sorted according to their frequency (decreasing). 

We observe that RoBERTa zero-prompt completely ignores all attributes with less than 150 instances. Furthermore, the only attributes that RoBERTa zero-prompt performs better are \textit{location, cleanness, temperature, size} and \textit{power}. Although for 4/5 of these attributes the difference in F1 score between the two models is marginal, the fact that 3/5 belong to the most frequent attributes (more than 1000 instances) influences the overall micro-F1.

\newpage
\subsection{Semantically Similar Attributes}

In \autoref{tab:synonyms} we show for every out-of-domain attribute, the most semantically similar in-domain attribute. This list contains only out-of-domain attributes that had a synonym from the in-domain group (Group Matched). This group was formed after manual inspection of the automatically generated synonym pairs.

\begin{table}[H]
\begin{tabular}{|ll|}
\hline
\textbf{Out-of-domain} & \textbf{In-domain}\\
\textbf{attribute} & \textbf{synonym/antonym}\\ \hline
activity                         & motion                     \\ 
angle                            & orientation                \\ 
area                             & shape                      \\ 
balance                          & weight                     \\ 
capacity                         & amount                     \\ 
\hline
consistency                      & stability                  \\ 
contents                         & composition                \\ 
direction                        & orientation                \\ 
flexibility                      & stability                  \\ 
granularity                      & composition                \\ 
\hline
height                           & length                     \\ 
hydration                        & wetness                    \\ 
intensity                        & brightness                 \\ 
quantity                         & amount                     \\ 
safety                           & speed                      \\ 
\hline
softness                         & hardness                   \\
tenseness                        & pressure                   \\
tension                          & pressure                   \\
thermal conductivity             & electric conductivity      \\
tightness                        & pressure                   \\
width                            & length                     \\ \hline

\end{tabular}
\caption{The most semantically similar in-domain attribute, each out-of-domain attribute.}
\label{tab:synonyms}
\end{table}

\begin{table*}[!ht]
\centering
\resizebox{1.7\columnwidth}{!}{
\begin{tabular}{lll}
\hline
\multicolumn{1}{c}{Semantic Cluster} & \multicolumn{1}{c}{In-domain Attributes}                                                                                            & \multicolumn{1}{c}{Out-of-domain Attributes}                                                                                                                \\ \hline
Spatial                              & \begin{tabular}[c]{@{}l@{}}location, volume, shape, orientation, \\ size, length, distance, organization\end{tabular}               & \begin{tabular}[c]{@{}l@{}}angle, direction, area, height, width, \\ pose, posture, spacial relation\end{tabular}                                           \\ \hline
Material                             & \begin{tabular}[c]{@{}l@{}}texture, electric conductivity, \\ thickness, hardness, strength, \\ pressure\end{tabular}               & \begin{tabular}[c]{@{}l@{}}tenseness, tension, tightness, \\ softness, material, flexibility, \\ thermal conductivity, density, \\ granularity\end{tabular} \\ \hline
Entity-Specific                      & \begin{tabular}[c]{@{}l@{}}cleanness, wetness, fullness, \\ ownership, openness, cost, \\ composition, coverage, focus\end{tabular} & \begin{tabular}[c]{@{}l@{}}contents, wholeness, capacity, \\ hydration, consumption, \\ documentation, emotional state, \\ pain, usage\end{tabular}         \\ \hline
Behavioral                           & \begin{tabular}[c]{@{}l@{}}knowledge, speed, motion, stability, \\ complexity, skill\end{tabular}                                   & \begin{tabular}[c]{@{}l@{}}activity, balance, consistency, \\ safety, familiarity, exposure, \\ viability, resistance\end{tabular}                          \\ \hline
Quantifier                           & amount                                                                                                                              & intensity, quantity, magnitude                                                                                                                              \\ \hline
Temporal                             & availability                                                                                                                        & age, life, existence, time                                                                                                                                  \\ \hline
Sensory Perception                   & visibility                                                                                                                          & \begin{tabular}[c]{@{}l@{}}color, taste, temperature, smell, sound, \\ appearance, weight, brightness\end{tabular}                                          \\ \hline
\end{tabular}}
\caption{Semantic clusters of attributes, both in-domain and out-of-domain.}
\label{tab:sem_clusters}
\end{table*}

\FloatBarrier

\subsection{Semantic Clusters of Attributes}
\label{app:sem_clusters}

Table \ref{tab:sem_clusters} shows the semantic clusters of attributes which are the result of agglomerative clustering and manually curation of in-domain and out-of-domain attributes. These clusters help better understand our attributes and performance based on their semantics. The clusters were used in  Section \ref{sec:sem_similarity}.

\subsection{OpenPI Real Examples}

Examples from out-of-domain with model predictions from the T5 $k$-attribute prompt and the RoBERTa single-attribute prompt models. In many instances the predicted attribute is correct, but the annotations fail to reflect this.

In \autoref{tab:examples}, we show some real instances that we used in our error analysis. Although for each instance all the out-of-domain attributes were queried, for brevity we only show attributes that were identified as changed by either model or by the annotations. We observe that in many of these examples the models predict attribute changes that are correct, despite not being captured by the annotations. Such cases are Example 2, Example 4 and Example 5, where the T5 $k$-attribute prompt correctly predicts attributes that were not identified by the annotators. These attributes are not necessarily related to the annotated attribute, such as \textit{width} and \textit{resistance} in Example 2, or \textit{hydration} and \textit{softness} in Example 4. However, some other instances may have predicted attributes that are closely related to the annotated attribute, as we see in Example 1, where \textit{posture} and \textit{angle} oftentimes change together. 

Our final observation from \autoref{tab:examples} is that the models are able to correctly predict attributes that require some common sense knowledge, which was not part of the provided context. For example, T5 $k$-attribute prompt predicts in Example 4 that \textit{soaking beans} implies that \textit{softness} changes, something that is not as an obvious conclusion as the change of \textit{hydration}. Even more, in Example 5 we observe that the model is able to understand the intent of the paragraph, which is to change the \textit{softness of lips}. These examples show that the T5 $k$-attribute prompt model is able to perform some degree of reasoning, even for predictions that were considered wrong due to missing annotations. 

\begin{table*}
\centering
\resizebox{2\columnwidth}{!}{
\begin{tabular}{ll}
\hline\\
\textbf{Example 1}\\[2pt]

\textbf{Context:} Begin by standing in Mountain Pose. Bend your right leg back and hold &\textbf{Entity:} person \\ on to the inside of your foot behind you with your right hand. & \textbf{Annotated Attributes:} balance \\[5pt]

\textbf{T5 $k$-attribute prompts:} Consider the following attributes: flexibility, angle, & \textbf{T5 $k$-attribute output:} posture, flexibility, \\  hydration, consumption. Which attribute changed for the person? & \hspace{115pt} angle, pose\\

\textbf{RoBERTa single-attribute prompts:} Is the flexibility of the person different? & \textbf{RoBERTa single-attribute output:} No\\
\hspace{168pt} Is the viability of the person different? & \hspace{162pt}Yes\\[5pt]
 \hline\\
  \textbf{Example 2}\\[2pt]
 \textbf{Context:} Cut off a corner of a yeast packet.&\textbf{Entity:} packet \\ & \textbf{Annotated Attributes:} resistance \\[8pt]

\textbf{T5 $k$-attribute prompts:} Consider the following attributes: contents, angle, & \textbf{T5 $k$-attribute output:} contents, width \\  width, resistance, softness. Which attribute changed for the packet? &\\

\textbf{RoBERTa single-attribute prompts:} Is the width of the packet different? & \textbf{RoBERTa single-attribute output:} Yes\\
\hspace{168pt} Is the resistance of the packet different? & \hspace{165pt}No \\[5pt]
 \hline\\
  \textbf{Example 3}\\[2pt]
 \textbf{Context:} Drink a glass of hot milk. &\textbf{Entity:} body \\ & \textbf{Annotated Attributes:} thermal conductivity \\[8pt]

\textbf{T5 $k$-attribute prompts:} Consider the following attributes: contents, hydration, & \textbf{T5 $k$-attribute output:} thermal conductivity \\  thermal conductivity. Which attribute changed for the body? &\\

\textbf{RoBERTa single-attribute prompts:} Is the thermal conductivity of the body different? & \textbf{RoBERTa single-attribute output:} No\\
\hspace{170pt}Is the hydration of the body different? & \hspace{160pt} Yes\\[5pt]
\hline\\
\textbf{Example 4}\\[2pt] 
 \textbf{Context:} Soak the dried beans and lentils overnight in a large bowl. &\textbf{Entity:} beans \\ & \textbf{Annotated Attributes:} hydration \\[8pt]

\textbf{T5 $k$-attribute prompts:} Consider the following attributes: softness, contents, & \textbf{T5 $k$-attribute output:} softness \\  granularity, hydration. Which attribute changed for the beans? &\\

\textbf{RoBERTa single-attribute prompts:} Is the hydration of the beans different? & \textbf{RoBERTa single-attribute output:} No\\
\hspace{170pt}Is the softness of the beans different? & \hspace{161pt} No\\[5pt]
\hline\\
 \textbf{Example 5}\\[2pt]
 \textbf{Context:} Take the honey and mix it with the sugar, then add in a little bit of Vaseline or &\textbf{Entity:} lips \\ petroleum jelly. When the mixture is all gritty, apply it on to your lips as you would with & \textbf{Annotated Attributes:} granularity \\ lip balm. Leave on the mixture for about one minute.\\ [8pt]

\textbf{T5 $k$-attribute prompts:} Consider the following attributes: softness, pain, & \textbf{T5 $k$-attribute output:} softness, pain \\  granularity. Which attribute changed for the lips? &\\

\textbf{RoBERTa single-attribute prompts:} Is the softness of the lips different? & \textbf{RoBERTa single-attribute output:} No\\
\hspace{170pt}Is the granularity of the lips different? & \hspace{161pt} No\\[5pt]
\hline
\end{tabular}
}

\caption{Examples from out-of-domain and model predictions for the T5 $k$-attribute prompt and the RoBERTa single-attribute prompt models.}
\label{tab:examples}
\end{table*}

\end{document}